\documentclass[10pt,twocolumn,letterpaper]{article}

\usepackage{icb}
\usepackage{times}
\usepackage{epsfig}
\usepackage{graphicx}
\usepackage{amsmath}
\usepackage{amssymb,adjustbox,ctable,threeparttable,booktabs}

\usepackage[pagebackref=true,breaklinks=true,letterpaper=true,colorlinks,bookmarks=false]{hyperref}

\icbfinalcopy 

\ificbfinal\fi
\begin{document}

\title{Robust Minutiae Extractor:\\ Integrating Deep Networks and Fingerprint Domain Knowledge}

\author{Dinh-Luan Nguyen, Kai Cao and Anil K. Jain\\
Michigan State University\\
East Lansing, Michigan, USA\\
{\tt\small nguye590@msu.edu, \{kaicao,jain\}@cse.msu.edu}
}

\maketitle
\thispagestyle{empty}

\begin{abstract}
	We propose a fully automatic minutiae extractor, called MinutiaeNet, based on deep neural networks with compact feature representation for fast comparison of minutiae sets. Specifically, first a network, called CoarseNet, estimates the minutiae score map and minutiae orientation based on convolutional neural network and fingerprint domain knowledge (enhanced image, orientation field, and segmentation map). Subsequently, another network, called FineNet, refines the candidate minutiae locations based on score map. We demonstrate the effectiveness of using the fingerprint domain knowledge together with the deep networks. Experimental results on both latent (NIST SD27) and plain (FVC 2004) public domain fingerprint datasets provide comprehensive empirical support for the merits of our method. Further, our method finds minutiae sets that are better in terms of precision and recall in comparison with state-of-the-art on these two datasets. Given the lack of annotated fingerprint datasets with minutiae ground truth, the proposed approach to robust minutiae detection will be useful to train network-based fingerprint matching algorithms as well as for evaluating fingerprint individuality at scale. MinutiaeNet is implemented in Tensorflow: \href{https://github.com/luannd/MinutiaeNet}{https://github.com/luannd/MinutiaeNet} \end{abstract}

\section{Introduction}
Automatic fingerprint recognition is one of the most widely studied topic in biometrics over the past $50$ years~\cite{jain201650}. One of the main challenges in fingerprint recognition is to increase the recognition accuracy, especially for latent fingerprints. Fingerprint comparison is primarily based on minutiae set comparison~\cite{zhao2007preprocessing, jain2007pores}. A number of hand-crafted approaches~\cite{jain1997line,zhao2007preprocessing} have been used to augment the minutiae with their attributes to improve the recognition accuracy. However, robust automatic fingerprint minutiae extraction, particularly for noisy fingerprint images, continues to be a bottleneck in fingerprint recognition systems.

\begin{figure}[!tbp]
\centering
\includegraphics[width=8cm]{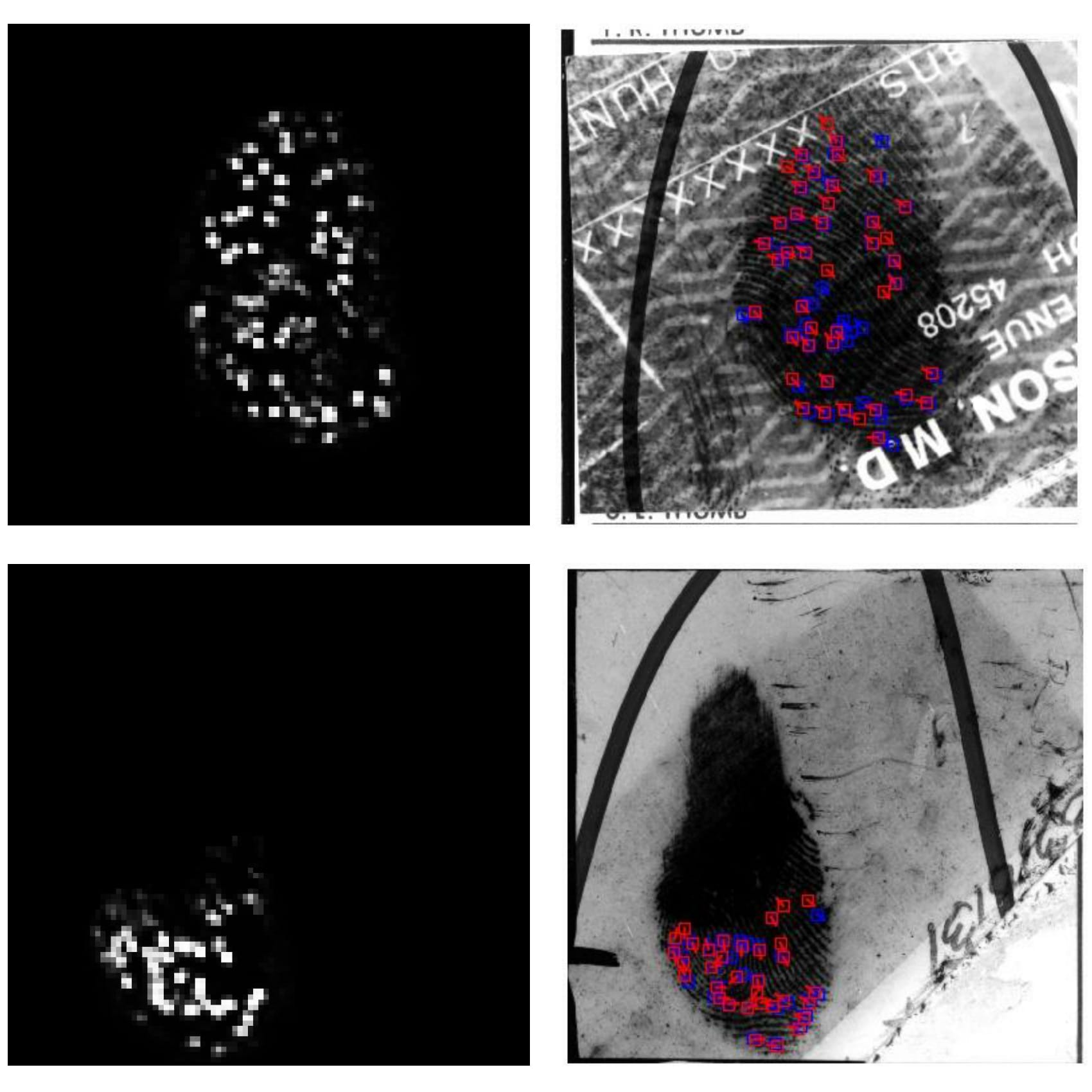}
\caption{Minutiae detection by the proposed approach on two latent fingerprint images ($\#7$ and $\#39$) from the NIST SD27 dataset~\cite{garris2000nist}. Left column: minutiae score maps obtained from the latent images shown in the right column. Right column: minutiae detected by the proposed framework (red) and ground truth minutiae (blue) overlaid on the latent image.}
\label{fig:Vis_intro}
\end{figure}

\begin{table*}[!tp]
	\centering
	\caption{Published network-based approaches for automatic minutiae extraction.}    
	\label{tab:existing_approach}
	\begin{small}
		\begin{threeparttable}
	\begin{adjustbox}{max width=\textwidth}
		\begin{tabular}{|c|c|c|c|c|c|}
			\specialrule{.15em}{.1em}{.1em}
			{\bfseries Study}& {\bfseries Method}& {\bfseries Training Data}& {\bfseries Testing Data}& {\bfseries Comments}& {\bfseries Performance Evaluation}\\
			\specialrule{.15em}{.1em}{.1em}
			{Sankaran \etal}&Sparse autoencoders&$132,560$ plain fingerprint images;&$129$ remaining latents&Sliding window; manual segmentation&Patch-based and minutia-based,\\
			{\cite{sankaran2014latent}}&for classification&$129$ images from NIST SD27&from NIST SD27& of latent fingerprints&metric and matching performance\tnote{(*)}\\
			
			\hline
			{Jiang \etal}&A combination of&$200$ live scan fingerprints& $100$ images from&Sliding window; hand-crafted dividing&Precision, recall,\\
			{\cite{jiang2016direct}}&JudgeNet and LocateNet&&private database&regions; no minutiae orientation information&and F1 score\\
			
			\hline
			{Tang \etal}&Fully convolutional&$4,205$ latent images from private&$129$ latents&Hard thresholds to cut off candidate regions;&Precision, recall, F1 score,\\
			{\cite{tang2017CNN}}&neural network&database and $129$ latents from NIST SD27&from NIST SD27&plain network&and matching performance\\
			
			\hline
			{Darlow \etal}&Convolutional network&$6,336$ images from&$1,584$ images from&Sliding window; hard threshold for candidate regions&Equal error rate and \\
			{\cite{Darlow2017DeepLearn}}&classifier &FVC 2000, 2002, and 2004 &FVC 2000, 2002, and 2004&(minutiae); separately estimated minutiae orientation&matching performance\\
			
			\hline
			{Tang \etal}&Unified network with&$8,000$ images from private&Set A of FVC 2004 and&Plain network; depends largely on the quality of&Precision, recall, and\\
			{\cite{tang2017FingerNet}}&domain knowledge&forensic latent database&NIST SD27&the enhancement and segmentation stages&matching performance\\
			\specialrule{.15em}{.1em}{.1em}
			{Proposed}&Domain knowledge with Residual&FVC 2002 with data augmentation&FVC 2004 ($3,200$ images) and&Residual network; automatic minutiae extractor&Precision, recall, and F1\\
			{approach}&learning based CoarseNet and&($8,000$ images in total)&NIST SD27 (all 258 latents)&utilizing domain knowledge; robust patch based&score under different location\\
			{}&inception-resnet based FineNet&&&minutiae classifier&and orientation thresholds\\
			\specialrule{.15em}{.1em}{.1em}
		\end{tabular}
		\end{adjustbox}
		\begin{tablenotes}
		\small
			\item[(*)] \footnotesize{different matchers were used in different studies and none of them were state of the art, i.e. top performing latent or slap matchers identified in the \\NIST evaluations. For this reason, we \emph{do not report} matching performance because otherwise it \emph{would not be a fair comparison} with previous studies.}
		\end{tablenotes}
		\end{threeparttable}
	\end{small}
\end{table*}

With rapid developments and success of deep learning techniques in a variety of applications in computer vision and pattern recognition~\cite{he2016deep, szegedy2017inception}, we are beginning to see network-based approaches being proposed for fingerprint recognition. Still, the prevailing methods of minutiae extraction primarily utilize fingerprint domain knowledge and handcrafted features. Typically, minutiae extraction and matching involves pre-processing stages such as ridge extraction and ridge thinning, followed by minutiae extraction~\cite{jain1997line, feng2008combining} and finally heuristics to define minutiae attributes. While such an approach works well for good quality fingerprint images, it provides inaccurate minutiae location and orientation for poor quality rolled/plain prints and, particularly for latent fingerprints. To overcome the noise in fingerprint images, Yoon \etal\cite{yoon2011latent} used Gabor filtering to calculate the reliability of extracted minutiae. Although this approach can work better than~\cite{jain1997line}, it also resulted in poor results with highly noisy images. Because these prevailing approaches are based on handcrafted methods or heuristics, they are only able to extract basic (or low level) features\footnote{Features such as edges, corners, etc.} of images. We believe learning based approaches using deep networks will have better ability to extract high level features\footnote{Abstract/semantic features retrieved from deep layers.} from low quality fingerprint images.

In this paper, we present a novel framework that exploits useful domain knowledge coded in the deep neural networks to overcome limitations of existing approaches to minutiae extraction. Figure \ref{fig:Vis_intro} visualizes results of the proposed framework on two latent fingerprints from the NIST SD27 dataset. 

Specifically, our proposed approach comprises of two networks, called CoarseNet and FineNet:

- \textbf{CoarseNet} is a residual learning~\cite{he2016deep} based convolutional neural network that takes a fingerprint image as initial input, and the corresponding enhanced image, segmentation map, and orientation field (computed by the early stages of CoarseNet) as secondary input to generate the minutiae score map. The minutiae orientation is also estimated by comparing with the fingerprint orientation.

- \textbf{FineNet} is a robust inception-resnet~\cite{szegedy2017inception} based minutiae classifier. It processes each candidate patch, a square region whose center is the candidate minutiae point, to refine the minutiae score map and approximate minutiae orientation by regression. Final minutiae are the classification results.

Deep learning approach has been used by other researchers for minutiae extraction (see Table \ref{tab:existing_approach}). But, our approach differs from published methods in the way we encode fingerprint domain knowledge in deep learning. Sankaran \etal~\cite{sankaran2014latent} classified the minutiae and non-minutiae patches by using sparse autoencoders. Jiang \etal~\cite{jiang2016direct} introduced a combination of two networks: JudgeNet for classifying minutiae patches, and LocateNet for locating precise minutiae location. While Jiang \etal use neural networks, their approach is very time-consuming due to use of sliding window to extract minutiae candidates. Another limitation of this approach is that it does not provide minutiae orientation information.

\begin{figure*}[!tbp]
\centering
\includegraphics[width=17.5cm]{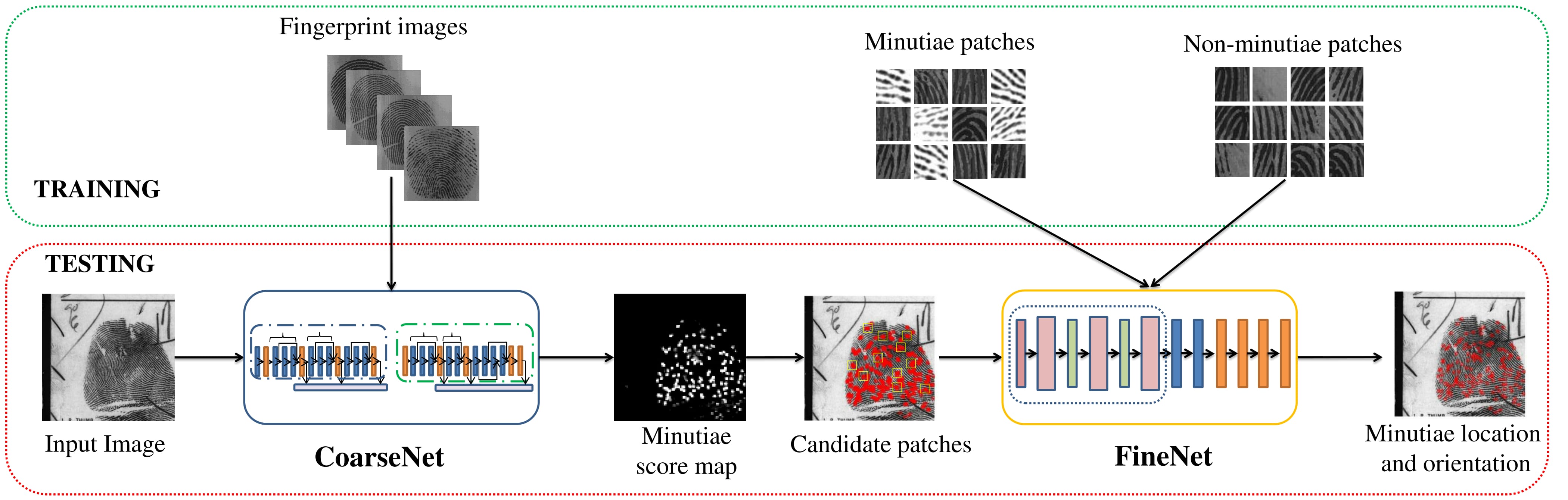}
\caption{Proposed automatic minutiae extraction architecture. While CoarseNet takes full fingerprint image as input, FineNet processes minutiae proposed patches output by CoarseNet.}
\label{fig:Full_architecture}
\end{figure*}

Tang \etal~\cite{tang2017CNN} utilized the idea of object detection to detect candidate minutiae patches, but it suffers from two major weaknesses: (i) hard threshold to delete the candidate patches, and (ii) the same network is used for both candidate generation and classification. By using sliding windows, Darlow \etal~\cite{Darlow2017DeepLearn} fed each pixel of the input fingerprint to a convolutional neural network, called MENet, to classify whether it corresponds to a minutia or not. It also suffers from time-consuming sliding windows as in~\cite{jiang2016direct}, and separate modules for minutiae location and orientation estimates. Tang \etal~\cite{tang2017FingerNet} proposed FingerNet that maps traditional minutiae extraction pipeline including orientation estimation, segmentation, enhancement, and extraction to a network with fixed weights. Although this approach is promising because it combines domain knowledge and deep network, it still uses plain \footnote{A series of stacked layers.} network architecture and hard threshold in non-maximum suppression \footnote{A post-processing algorithm that merges all detections belonging to the same object.}. Finally, the accuracy of FingerNet depends largely on the quality of the enhanced and segmentation stage while ignoring texture information in the ridge pattern.

In summary, the published approaches suffer from using sliding windows to process each pixel in input images, setting hard threshold in post-processing step, and using plain convolutional neural network to classify candidate regions. Furthermore, the evaluation process in these studies is not consistent in terms of defining ``correct'' minutiae.

The contributions of our approach are as follows:
\vspace{-2mm}
\begin{itemize}
\item A network-based \emph{automatic minutiae extractor} utilizing domain knowledge is proposed to provide reliable minutiae location and orientation without a hard threshold or fine tuning.
\vspace{-1mm}
\item A \emph{robust patch based minutiae classifier} that significantly boosts the precision and recall of candidate patches. This can be used as a robust minutiae extractor with compact embedding of minutiae features.
\vspace{-1mm}
\item A \emph{non-maximum suppression} is proposed to get precise locations for candidate patches. Experimental evaluations on FVC 2004~\cite{maio2004fvc2004} and NIST SD27~\cite{garris2000nist} show that the proposed approach is superior to published approaches in terms of precision, recall, and F1 score values.
\end{itemize}


\section{Proposed framework}
\label{sec:sec_Proposed_framework}
Our minutiae extraction framework has two modules: (i) residual learning based convolutional neural network, called CoarseNet that generates candidate patches containing minutiae from input fingerprint image; (ii) inception-resnet based network architecture, called FineNet which is a strong minutiae classifier that classifies the candidate patches output by CoarseNet. These two networks also provide minutiae location and orientation information as outputs. Figure \ref{fig:Full_architecture} describes the complete network architecture for automatic minutiae location and orientation for an input fingerprint image. Section \ref{sec:sec_CoarseNet} presents the architecture of CoarseNet. In Section \ref{sec:sec_FineNet}, we introduce FineNet with details on training to make it a strong classifier.

\begin{figure*}[tbp]
\centering
\includegraphics[height=5.2cm]{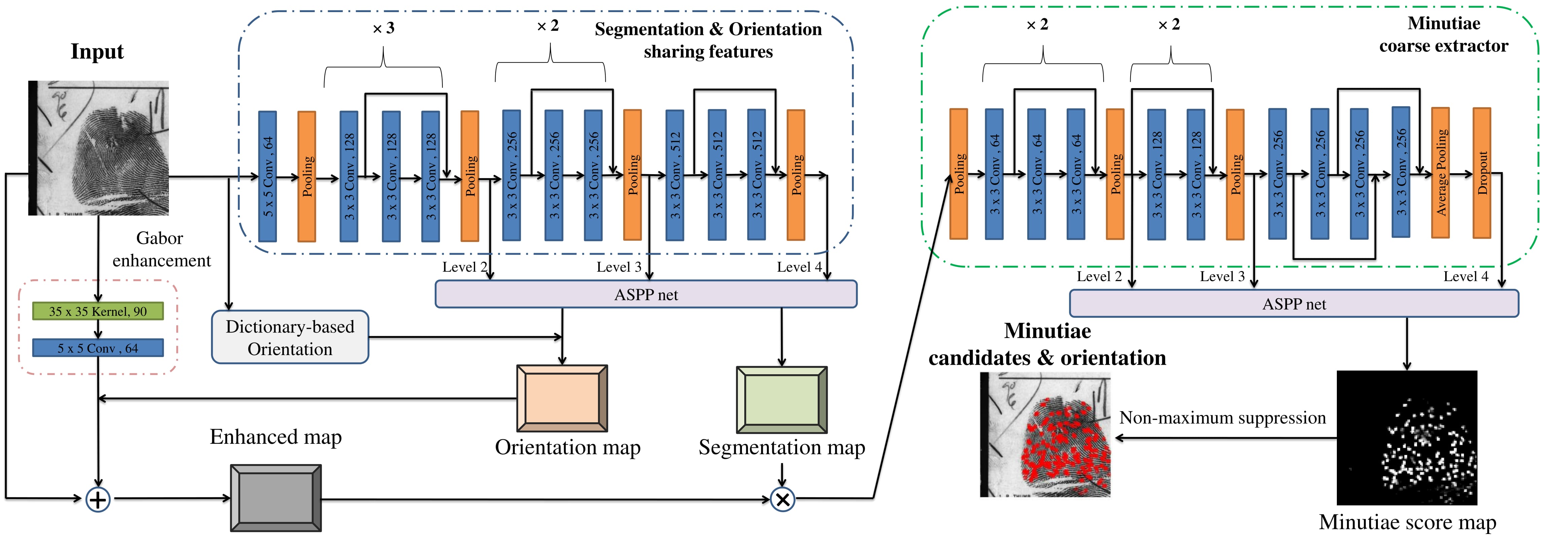}
\caption{CoarseNet architecture.}
\label{fig:CoarseNet}
\end{figure*}

\subsection{CoarseNet for minutiae extraction}
\label{sec:sec_CoarseNet}
We adopt the idea of combining domain knowledge and deep representation of neural networks in~\cite{tang2017FingerNet} to boost the minutiae detection accuracy. In essence, we utilize the automatically extracted segmentation map, enhanced image, and orientation map as complementary information to the input fingerprint image. The goal of CoarseNet is not to produce the segmentation map or enhanced image or orientation map. They are just the byproducts of the network. However, these byproducts as fingerprint domain knowledge must be reliable to get robust minutiae score map. Because Tang \etal~\cite{tang2017FingerNet} proposed an end-to-end unified network that maps handcrafted features to network based architecture, we use this as a baseline for our CoarseNet.

\subsubsection{Segmentation and orientation feature sharing}
\label{sec:sec_Segmentation_and_Orientation}
Adding more layers in the deep network with the hope of increasing accuracy might lead to the exploding or vanishing gradients problem. From the success of residual learning~\cite{he2016deep}, we use residual instead of just plain stacked convolutional layers in our network to make it more powerful. Figure \ref{fig:CoarseNet} shows the detailed architecture of the network. 

Unlike existing works using plain convolutional neural network~\cite{tang2017CNN, tang2017FingerNet} or sliding window~\cite{sankaran2014latent, Darlow2017DeepLearn} to process each patch with fixed size and stride, we use a deeper residual learning based network with more pooling layers to scale down the region patch. Specifically, we get the output after the $2^{nd}$, $3^{rd}$, and $4^{th}$ pooling layer to feed to an ASPP network~\cite{chen2016deeplab} with corresponding rates for multiscale segmentation. This ensures the output has the same size as input without a loss of  information when upsampling the score map.

By using four pooling layers, each pixel in the $j^{th}$ feature map, called level $j$, corresponds to a region $2^j \times 2^j$ in the original input. Result layers at level $4$ and $3$ will be tested as coarse estimates while the level $2$ serves as fine estimation.

\textbf{Segmentation map.}
Image segmentation and fingerprint orientation estimation share the same convolutional layers. Thus, by applying multi-level approach mentioned above, we get probability maps of each level-corresponding region in input image. For instance, to get finer-detailed segmentation for each region level $j_l$, we continue to process probability map of region level $j_l/2$.

\textbf{Orientation map.}
To get complete minutiae information from context, we adopt the fusion idea of Cao \etal~\cite{cao2017automated}. We fuse the results of Dictionary-based method~\cite{yang2014localized} with our orientation results from CoarseNet. Because~\cite{yang2014localized} uses a hand-crafted approach, we set the fusion weight ratio of its output with our network-based approach as 1:3.
\vspace{-2mm}
\subsubsection{Candidate generation}
The input fingerprint image might contain large amounts of noise. So, without using domain knowledge we may not be able to identify prominent fingerprint features. The domain knowledge comprises of four things: raw input image, enhanced image, orientation map, and segmentation map. In the Gabor image enhancement module, we take the average of filtered image and the orientation map for ridge flow estimation. To emphasize texture information in fingerprints, we stack the original input image with the output of enhancement module to obtain the final enhancement map. To remove spurious noises, we apply segmentation map on the enhancement map and use it as input to coarse minutiae extractor module.

\begin{figure}[!bp]
\centering
\includegraphics[width=6.5cm]{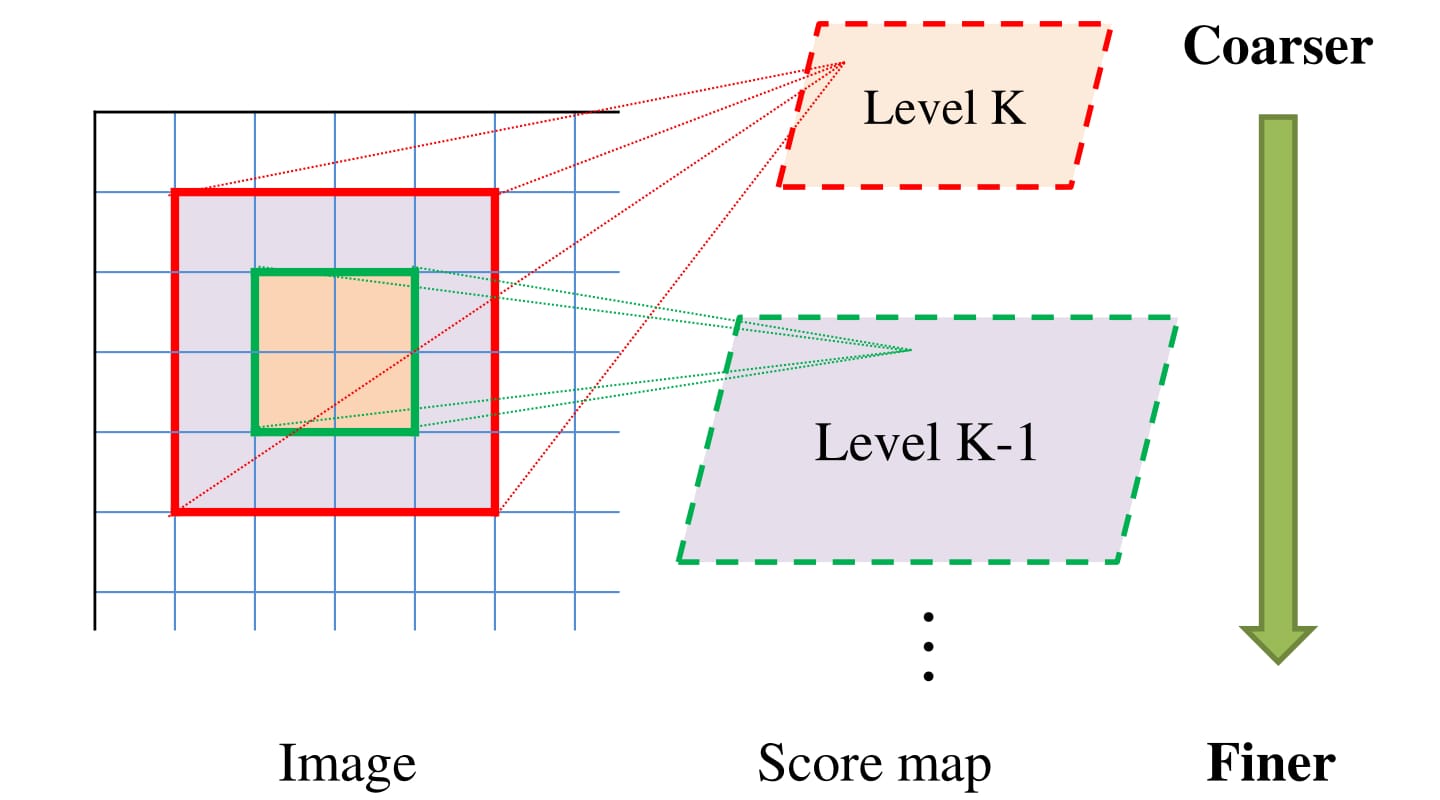}
\caption{Candidate patch processing map. Lower the level of score map, more detail it provides.}
\label{fig:candidate_process_map}
\end{figure}

To obtain the precise location of minutiae, each level of residual net is fused to get the final minutiae score map with size $h/16 \times w/16$, where $h$ and $w$ are the height and width of the input image. Figure \ref{fig:candidate_process_map} shows the details of processing score map. To reduce the processing time, we use score map at level $4$ as a coarse location. To get precise location, lower level score maps are used.
\vspace{-2mm}
\subsubsection{Non-maximum suppression}
\label{sec:sec_NMS}
Using non-maximum suppression to reduce the number of candidates is common in object detection~\cite{girshick2015deformable, ren2017faster}. Some of the candidate regions are deleted to get a reliable minutiae score map by setting a hard threshold~\cite{Darlow2017DeepLearn} or using heuristics~\cite{tang2017CNN, tang2017FingerNet}. However, a hard threshold can also suppress valid minutiae locations. A commonly used  heuristics is to sort the candidate scores in ascending order. The $L^2$ distance between pairwise candidates is calculated with hard thresholds for distance and orientation. By iteratively comparing each candidate with the rest in the candidate list, only the candidate with higher score and score above the thresholds is kept. However, this approach fails when two minutiae are near each other and the inter-minutiae distance is below the hard thresholds.

Since each score in the minutiae map corresponds to a specific region in the input image, we propose to use the intersection over union strategy. Specifically, after sorting the scores of the candidate list, we keep high score candidates while ignoring the lower scoring candidates with at least $50\%$ overlap with the candidates already selected.
\vspace{-4mm}
\subsubsection{Training data for CoarseNet}
\label{sec:lack_data}
Given the lack of datasets with ground truth, we use the approach in Tang \etal~\cite{tang2017FingerNet} to generate weak labels for training the segmentation and orientation module. The coarse minutiae extractor module uses minutiae location and minutiae orientation ground truth provided in the two datasets. We also use data augmentation techniques as mentioned in Section \ref{sec:sec_Experiments}.

\subsection{FineNet}
\label{sec:sec_FineNet}
Extracting minutiae based on candidate patches is not adequate. Although CoarseNet is reliable, it still fails to detect true minutiae or detects spurious minutiae. This can lead to poor performance in fingerprint matching. This motivates our use of FineNet - a minutiae classifier from generated candidate patches. FineNet takes candidates from the output of CoarseNet as input to decide whether the region $10 \times 10$ in the center of the corresponding patch has a valid minutia or not.
\vspace{-4mm}
\subsubsection{FineNet architecture}
\label{sec:subsec_FineNet_architecture}
\begin{figure}[!bp]
\centering
\includegraphics[width=6cm]{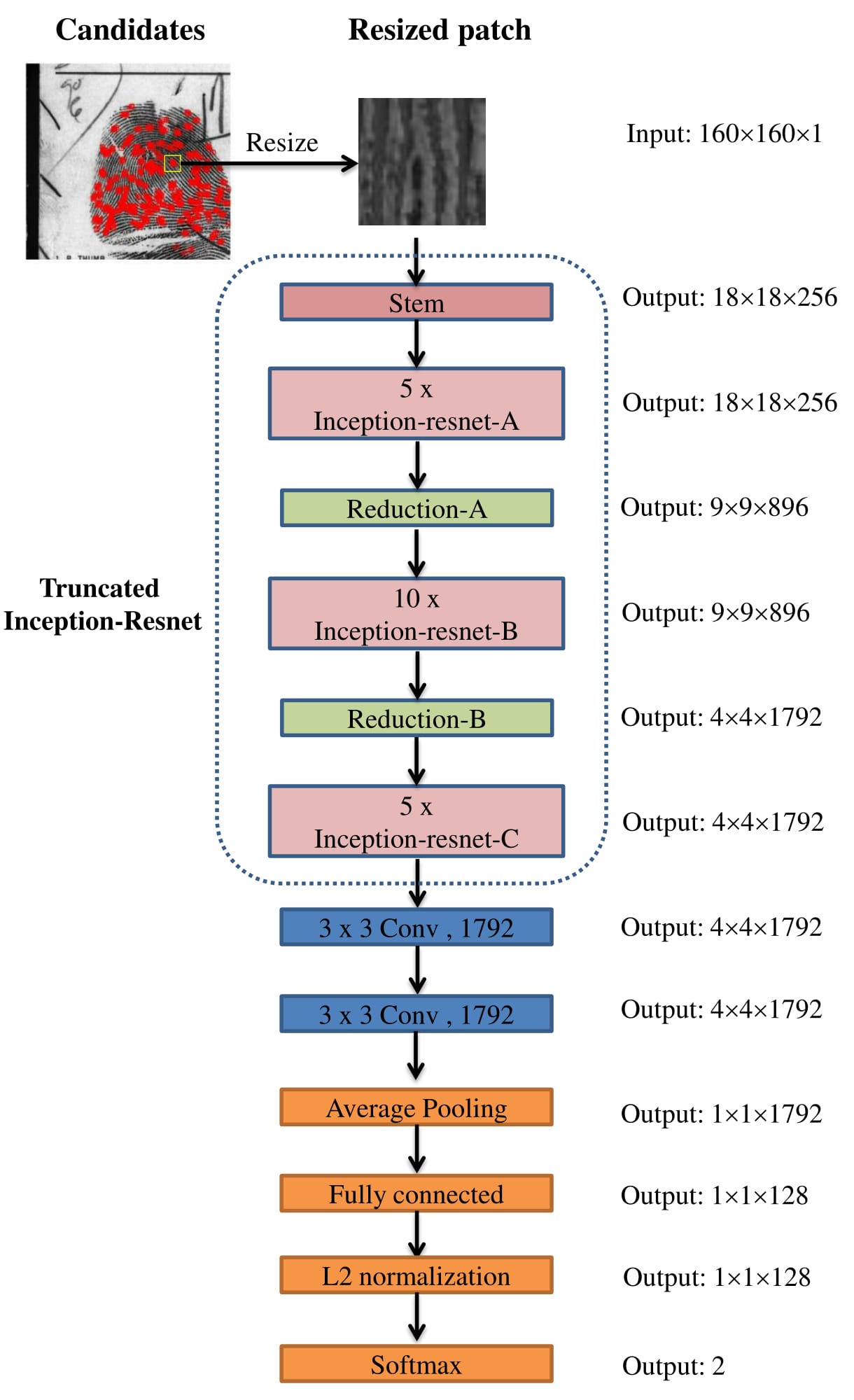}
\caption{FineNet architecture. For details of the Inception-Resnet v1 arichitecture block, we refer the readers to~\cite{szegedy2017inception}.}
\label{fig:FineNet}
\end{figure}

Figure \ref{fig:FineNet} describes the architecture of FineNet. As mentioned in Section \ref{sec:sec_Segmentation_and_Orientation}, we use the Inception-Resnet v1 architecture as a core network in FineNet. 

\begin{table*}[!tbhp]
	\centering
	\caption{Comparison of different methods for minutiae extraction on FVC 2004 and NIST SD27 datasets. Note that~\cite{sankaran2014latent, tang2017CNN} reported their results only on subsets of FVC 2004 and NIST SD27 as mentioned in Table \ref{tab:existing_approach}. ``\_'' means the authors neither provided these results in their paper nor made their code available. $\mathcal{D}$ and $\mathcal{O}$ are parameters defined in Eq. (\ref{eq:D_O}).}    
	\label{tab:compare_configuration}
	\begin{small}
		\begin{adjustbox}{max width=\textwidth}
		\begin{tabular}{|c|c|c|c|c|c|c|c|c|c|c|}
			\hline
			{\bfseries Dataset}&{\bfseries Methods}& \multicolumn{3}{c|}{\bfseries Setting 1 $(\mathcal{D}=8,\mathcal{O}=10)$}& \multicolumn{3}{c|}{\bfseries Setting 2 $(\mathcal{D}=12,\mathcal{O}=20)$}&\multicolumn{3}{c|}{\bfseries Setting 3 $(\mathcal{D}=16,\mathcal{O}=30)$}\\
			\cline{3-11}
			&{}&{\bfseries Precision}& {\bfseries Recall}& {\bfseries F1 score}&{\bfseries Precision}& {\bfseries Recall}& {\bfseries F1 score}& {\bfseries Precision}& {\bfseries Recall}& {\bfseries F1 score}\\
			\hline
			{}&{MINDTCT~\cite{ko2007user}}&$8.3\%$&$14.7\%$&$0.106$&$10.0\%$&$16.4\%$&$0.124$&$11.2\%$&$18.9\%$&$0.141$\\
			{}&{VeriFinger~\cite{verifinger2010neuro}}&$3.6\%$&$40.1\%$&$0.066$&$5.3\%$&$47.9\%$&$0.095$&$7.6\%$&$58.3\%$&$0.134$\\
			{}&{Gao \etal~\cite{gao2010novel}}&\_ &\_ &\_ &\_ &\_ &\_ &$23.5\%$&$8.7\%$&$0.127$\\
			{NIST SD27}&{Sankaran \etal~\cite{sankaran2014latent}}&\_ &\_ &\_ &\_ &\_ &\_ &$26.4\%$&$63.1\%$&$0.372$\\
			{}&{Tang \etal~\cite{tang2017CNN}}&\_ &\_ &\_ &\_ &\_ &\_ &$53.0\%$&$53.4\%$&$0.532$\\
			{}&{FingerNet~\cite{tang2017FingerNet}}&$53.2\%$&$49.5\%$&$0.513$&$58.0\%$&$58.1\%$&$0.58$&$63.0\%$&$63.2\%$&$0.631$\\
			\cline{2-11}
			{}&{\bfseries Proposed method}&{\bfseries 69.2\%}&{\bfseries 67.7\%}&{\bfseries 0.684}&{\bfseries 70.5\%}&{\bfseries 72.3\%}&{\bfseries 0.714}&{\bfseries 71.2\%}&{\bfseries 75.7\%}&{\bfseries 0.734}\\
			\hline
			
			{}&{MINDTCT~\cite{ko2007user}}&$30.8\%$&$64.3\%$&$0.416$&$37.7\%$&$72.1\%$&$0.495$&$42.1\%$&$79.8\%$&$0.551$\\
			{}&{VeriFinger~\cite{verifinger2010neuro}}&$39.8\%$&$69.2\%$&$0.505$&$45.6\%$&$77.5\%$&$0.574$&$51.8\%$&$81.9\%$&$0.635$\\
			{FVC 2004}&{Gao \etal~\cite{gao2010novel}}&\_ &\_ &\_ &\_ &\_ &\_ &$48.8\%$&$82.7\%$&$0.614$\\
			{}&{FingerNet~\cite{tang2017FingerNet}}&$68.7\%$&$62.1\%$&$0.643$&$72.9\%$&$70.4\%$&$0.716$&$76.0\%$&$80.0\%$&$0.779$\\
			\cline{2-11}
			{}&{\bfseries Proposed method}&{\bfseries 79.0\%}&{\bfseries 80.1\%}&{\bfseries 0.795}&{\bfseries 83.6\%}&{\bfseries 83.9\%}&{\bfseries 0.837}&{\bfseries 85.9\%}&{\bfseries 84.8\%}&{\bfseries 0.853}\\
			\hline
			
		\end{tabular}
		\end{adjustbox}
	\end{small}
\end{table*}

For FineNet training, we extract an equal number of $t_1 \times t_1$ sized minutiae and non-minutiae patches with $t_1 = 45$. FineNet determines the whether the $10 \times 10$ pixel region in the center of each patch contains a valid minutia or not. The candidate patches are resized into $t_2 \times t_2$ pixels that feed to FineNet. Based on the observation that the original input image size (without rescaling) is not large in comparison with images for object classification, too much up scaling the image can cause blurring the tiny details, and too small an input image size is not sufficient for network with complex architecture, we choose $t_2 = 160$ pixels.

Training data for FineNet is extracted from the input gray scale images where minutiae data are based on the ground truth minutiae location and non-minutiae ones are from random sampling with the center regions do not contain partial or fully minutiae location. To make the network more robust, we use some small techniques like Batch Normalization~\cite{ioffe2015batch}, rotation, flipping, scaled augmentation~\cite{simonyan2014very}, and bounding blurring~\cite{jiang2016direct} as pre-processing stage. 
\vspace{-2mm}
\subsubsection{Losses and implementation details}
\textbf{Intra-Inter class losses.}
Because input captured fingerprint image is not always in the ideal condition, it might be easily affected by distortion, finger movement, or quality (wet/dry) of finger. Thus, the variation of minutiae shapes and surrounding ridges can affect the accuracy of the classifier. To handle this situation and make the FineNet more robust with intra-class variations, we use Center Loss~\cite{wen2016discriminative} as a complementary of softmax loss and minutiae orientation loss. While softmax loss tends to push away features of different classes, center loss tends to pull features in the same class closer. Let $\mathcal{L}$, $\mathcal{L_C}$, $\mathcal{L_S}$, $\mathcal{L_O}$ be the total loss, center loss, sofmax loss, and orientation loss, the total loss for training is calculated as follows:
\vspace{-2mm}
\begin{equation}
\mathcal{L} = \alpha \mathcal{L_C} + (1-\alpha)\mathcal{L_S} + \beta \mathcal{L_O}
\label{eq:loss_FineNet}
\end{equation}
where we set $\alpha = 0.5$ to balance between intra-class (center) and inter-class (sofmax) loss and $\beta = 2$ to emphasize the importance of precision of minutiae orientation.

\textbf{Parameter settings.}
As mentioned in Section \ref{sec:subsec_FineNet_architecture}, fingerprint patches are input to FineNet where the patch size is $160 \times 160$. To ensure our network can handle distortion in input image, we apply scaled augmentation~\cite{simonyan2014very}, random cropping, and brightness adjustment. Horizontal and vertical flip with pixel mean subtraction are also adopted. We randomly initialize all variables by using a Gaussian distribution $N(0, 0.01)$. Batch size is set to $100$. We use schedule learning rate after particular iterations. Specifically, we set it as $0.01$ in the beginning and reduce $10$ times after $50K$ iterations. To prevent vanishing gradient problem, we set the maximum epoch to $200K$. We use $0.0004$ as the value for momentum and weight decay is $0.9$.
\vspace{-2mm}
\section{Experimental results}
\label{sec:sec_Experiments}
We evaluate our method on two different datasets with different characteristics under different settings of parameters $\mathcal{D}$ and $\mathcal{O}$ (see Eq. (\ref{eq:D_O})). We also visualize examples of score maps with correct and incorrect minutiae extractions in Figure \ref{fig:visualize}. All experiments were implemented in Tensorflow and ran on Nvidia GTX GeForce. 
\vspace{-2mm}
\subsection{Datasets}
We use FVC 2002 dataset~\cite{maio2002fvc2002} with data augmentation consisting of $3,200$ plain fingerprint images for training. To compensate for the lack of a large scale dataset for training, we distort the input images in $x$ and $y$ coordinates in the spirit of hard-training with non-ideal input fingerprint images. Furthermore, we also apply additive random noise to the input images. Thus, for training CoarseNet, we have an augmented dataset of $8,000$ images. To obtain data for training FineNet, we extract $45 \times 45$ pixel patches from these $8K$ training images for CoarseNet whose center is a ground truth minutia point. For non-minutiae patches, we randomly extract patches with the criteria that the $10 \times 10$ center of each patch does not contain any minutia. Thus, we collect around $100K$ minutia and non-minutia patches for training FineNet.

As mentioned in Section \ref{sec:lack_data}, we use FingerNet~\cite{tang2017FingerNet} to generate labels for domain knowledge groundtruth. Besides, we manually correct segmentation grountruth results from FingerNet to ensure better learning for CoarseNet.
\vspace{-1mm}
\subsection{Evaluation}

To demonstrate the robustness of our framework, we compare our results with published approaches on FVC 2004~\cite{maio2004fvc2004}\footnote{We obtained the groundtruth from \cite{kayaoglu2013standard}} and NIST SD27~\cite{garris2000nist} datasets under different criteria of distance and orientation thresholds. Let the tuples $(l_p,o_p)$ and $(l_{gt},o_{gt})$ be the location coordinates and orientation values of predicted and ground truth minutia. The predicted minutia is called true if it satisfies the following constrains:
\vspace{-2mm}
\begin{equation}
\label{eq:D_O}
\begin{cases}
{\lVert \mathbf{l_p-l_{gt}} \rVert}_2 \leq \mathcal{D}   \\
{\lVert \mathbf{o_p-o_{gt}} \rVert}_1 \leq \mathcal{O}
\end{cases}
\end{equation}

where $\mathcal{D}$ and $\mathcal{O}$ are the thresholds in pixels and degrees, respectively.
Specifically, we set the range of distances between detected and ground truth minutiaes from $8$ to $16$ pixels (in location) and $10$ to $30$ degree (in orientation) with default threshold value ($0.5$). We choose these settings to demonstrate the robust and precise results from the proposed approach while published works degrade rather quickly.

\begin{figure*}[tbp]
\centering
\includegraphics[width=16.5cm]{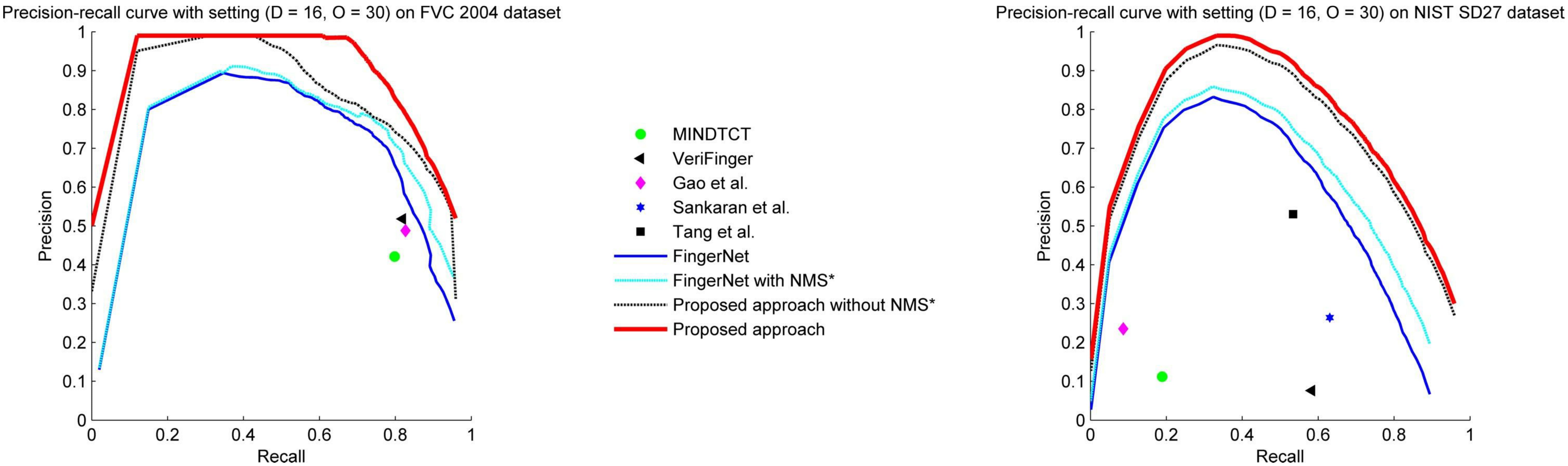}
\caption{Precision-Recall curves on FVC 2004 (left) and NIST SD27 (right) datasets with published approaches in Setting 3.}
\label{fig:Graph_curve}
\end{figure*}
	
Table \ref{tab:compare_configuration} shows the precision and recall comparisons of different approaches to minutiae extraction. MINDTCT~\cite{ko2007user} is the open source NIST Biometric Image Software. VeriFinger~\cite{verifinger2010neuro} is a commercial SDK for minutiae extraction and matching. Since Gao \etal~\cite{gao2010novel} did not release their code in public domain, we report their results on NIST SD27 and FVC 2004 database from~\cite{tang2017FingerNet}. Darlow \etal~\cite{Darlow2017DeepLearn} use only a subset of the FVC dataset for training and the rest for testing, we do not include in our evaluation.

Table \ref{tab:compare_configuration} shows that the proposed method outperforms state-of-the-art techniques under all settings of parameters (thresholds) $\mathcal{D}$ and $\mathcal{O}$ for both FVC 2004 and NIST SD27. Our results also reveal that by using only rolled/plain fingerprint images for training, our framework can work pretty well for detecting minutiae in latents.

\begin{table}[!tp]
	\centering
	\caption{The importance of non-maximum suppression in our framework. $NMS$ and $NMS^*$ denote the non-maximum approaches of FingerNet and the proposed approach, respectively.}    
	\label{tab:NMS}
	\begin{small}
	\begin{adjustbox}{max width=\textwidth}
		\begin{tabular}{|c|c|c|c|}
			\hline
			\cline{2-3}
			{\bfseries Configuration}& {\bfseries Precision}& {\bfseries Recall}& {\bfseries F1 score}\\
			\hline
			{FingerNet + $NMS$}	&$63.0\%$&$63.2\%$&$0.631$\\
			{FingerNet + $NMS^*$}	&$65.2\%$&$65.4\%$&$0.653$\\
			{Proposed method + $NMS$}	&	$69.4\%$&	$73.5\%$&$0.714$\\
			{\bfseries Proposed method + $NMS^*$}	&{\bfseries 71.2\%}&{\bfseries 75.7\%}&{\bfseries 0.734}\\
			\hline
		\end{tabular}
		\end{adjustbox}
	\end{small}
\end{table}

Table \ref{tab:NMS} shows a comparison between using and not using our proposed non-maximum suppression method on the NIST SD27 dataset with setting $3$ in Table \ref{tab:compare_configuration}. Because non-maximum suppression is a post processing step, it helps improve precision, recall and F1 values.

To make a complete comparison (at all the operating points) with published methods, we present the precision-recall curves in Figure \ref{fig:Graph_curve}. The proposed approach surpasses all published works on both FVC 2004 and NIST SD27 datasets.

\begin{figure*}[!tbp]
\centering
\includegraphics[width=15cm]{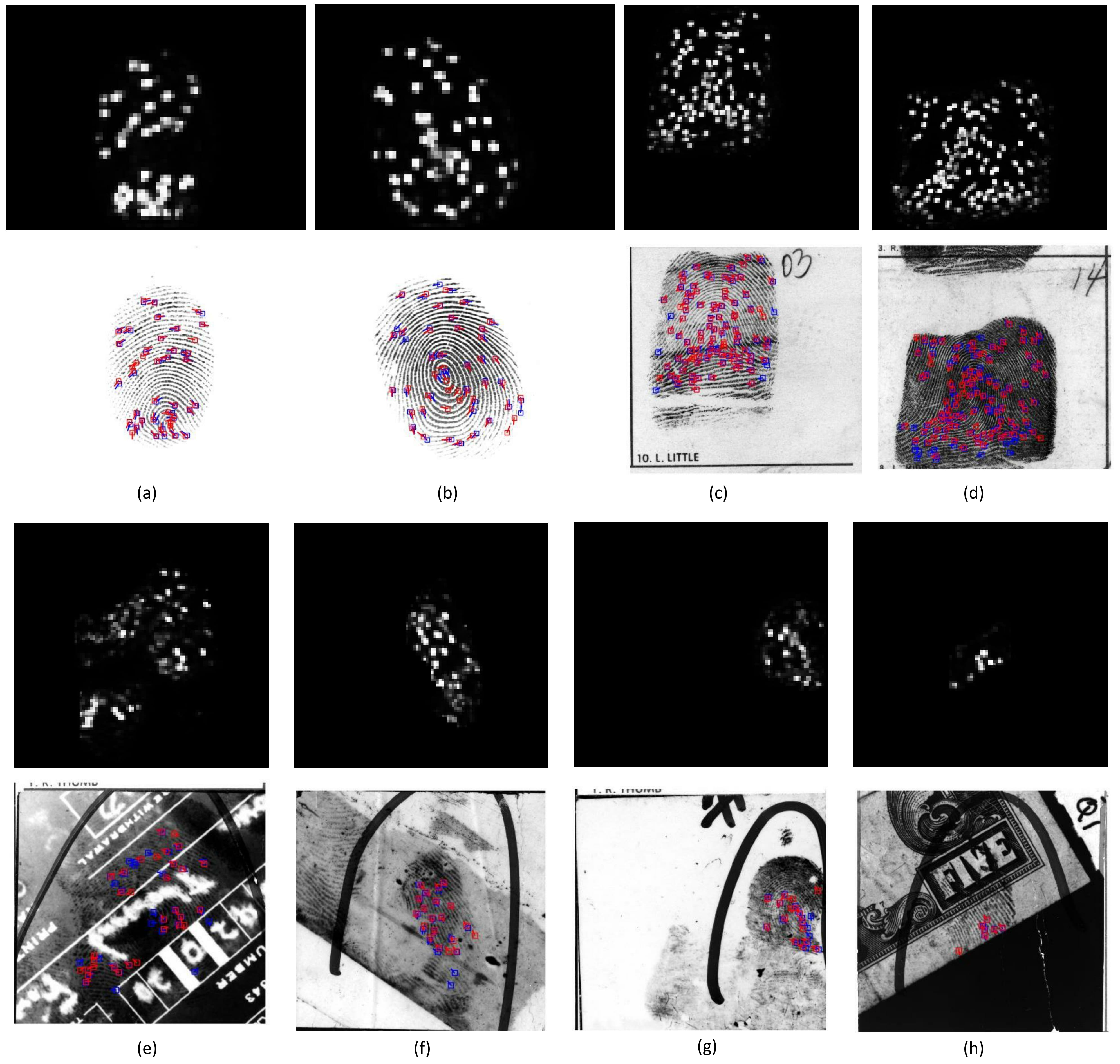}
\caption{Visualizing minutiae extraction results. From top to bottom in each column: score maps, and minutiae extraction results overlaid on fingerprint images. (a)-(b): two plain images from FVC 2004; (c)-(d): rolled (reference) fingerprints from NIST SD27; (e)-(h): latent fingerprint images from NIST SD27.}
\label{fig:visualize}
\end{figure*}

Figure \ref{fig:visualize} shows the minutiae extraction results on both FVC 2004 and NIST SD27 datasets with different quality images. Our framework works well in difficult situations such as noisy background or dry fingerprints. However, there are some cases where the proposed framework either misses the true minutiae or extracts spurious minutiae. For the FVC 2004 dataset and rolled fingerprints from NIST SD27 dataset, we obtain results that are close to the ground truth minutiae. However, some minutiae points are wrongly detected (image a) because of the discontinuity of ridges or missed detections (image c) because the location of minutiae is near the fingerprint edge. For the latent fingerprints from NIST SD27 dataset, besides the correctly extracted minutiae, the proposed method is sensitive to severe background noise (image e) and poor latent fingerprint quality (image g). The run time per image is around 1.5 seconds for NIST SD27 and 1.2 seconds for FVC 2004 on Nvidia GTX GeForce.

\section{Conclusions}
\label{sec:sec_Conclusion}
\vspace{-2mm}
We have presented two network architectures for automatic and robust fingerprint minutiae extraction that fuse fingerprint domain knowledge and deep network representation:

- {\bfseries CoarseNet}: an automatic robust minutiae extractor that provides candidate minutiae location and orientation without a hard threshold or fine tuning.

- {\bfseries FineNet}: a strong patch based classifier that accelerates the reliability of candidates from CoarseNet to get final results.

A non-maximum suppression is proposed as a post processing step to boost the performance of the whole framework. We also reveal the impact of residual learning on minutiae extraction in the latent fingerprint dataset despite using only plain fingerprint images for training. Our experimental results show that the proposed framework is robust and achieves superior performance in terms of precision, recall and F1 values over published state-of-the-art on both benchmark datasets, namely FVC 2004 and NIST SD27.

The proposed framework can be further improved by (i) using larger training set for network training that includes latent images, (ii) constructing context descriptor to exploit the region surrounding minutiae, (iii) improving processing time, and (iv) unifying minutiae extractor into an end-to-end fingerprint matching framework.

{\footnotesize
\bibliographystyle{ieee}
\bibliography{ICB_ref}
}

\end{document}